\documentclass{article} 
\usepackage{nips14submit_e,times}
\usepackage{url}
\usepackage{amsmath, algorithm}
\usepackage{algpseudocode}
\usepackage[authoryear,round]{natbib}
\usepackage{booktabs}

\DeclareMathOperator*{\argmax}{arg\,max}

\title{First-Pass Large Vocabulary Continuous Speech Recognition using Bi-Directional Recurrent DNNs}

\author{
Awni Y. Hannun \\
Computer Science Department\\
Stanford University\\
Stanford, CA 94305 \\
\texttt{awni@cs.stanford.edu} \\
\And
Andrew L. Maas \\
Computer Science Department\\
Stanford University\\
Stanford, CA 94305 \\
\texttt{amaas@cs.stanford.edu} \\
\AND
Daniel Jurafsky \\
Linguistics Department\\
Stanford University\\
Stanford, CA 94305 \\
\texttt{jurafsky@stanford.edu} \\
\And
Andrew Y. Ng \\
Computer Science Department\\
Stanford University\\
Stanford, CA 94305 \\
\texttt{ang@cs.stanford.edu}
}

\nipsfinalcopy 

\begin{document}

\maketitle

\begin{abstract}
We present a method to perform first-pass large vocabulary continuous
speech recognition using only a neural network and language
model. Deep neural network acoustic models are now commonplace in
HMM-based speech recognition systems, but building such systems is a
complex, domain-specific task.  Recent work demonstrated the
feasibility of discarding the HMM sequence modeling framework by
directly predicting transcript text from audio. This paper extends
this approach in two ways. First, we demonstrate that a
straightforward recurrent neural network architecture can achieve a
high level of accuracy. Second, we propose and evaluate a modified
prefix-search decoding algorithm. This approach to decoding enables
first-pass speech recognition with a language model, completely
unaided by the cumbersome infrastructure of HMM-based
systems. Experiments on the Wall Street Journal corpus demonstrate
fairly competitive word error rates, and the importance of
bi-directional network recurrence.
\end{abstract}

\section{Introduction}\label{sec:introduction}
Modern large vocabulary continuous speech recognition (LVCSR) systems
are complex and difficult to modify. Much of this complexity stems
from the paradigm of modeling words as sequences of sub-phonetic
states with hidden Markov models (HMMs). HMM-based systems require
carefully-designed training recipes to construct consecutively more
complex HMM recognizers. The overall difficulty of building,
understanding, and modifying HMM-based LVCSR systems has limited
progress in speech recognition and isolated it from many advances in
related fields. 

Recently \citet{Graves2014} demonstrated an HMM-free approach to
training a speech recognizer which uses a neural network to directly
predict transcript characters given the audio of an utterance. This
approach discards many of the assumptions present in modern HMM-based
LVCSR systems in favor of treating speech recognition as a direct
sequence transduction problem. The approach trains a neural network
using the connectionist temporal classification (CTC) loss function,
which amounts to maximizing the likelihood of an output sequence by
efficiently summing over all possible input-output sequence
alignments. Using CTC the authors were able to train a neural network
to predict the character sequence of test utterances with a character
error rate (CER) under 10\% on the Wall Street Journal LVCSR
corpus. While impressive in its own right, these results are not yet
competitive with existing HMM-based systems in terms of word error
rate (WER). Good word-level performance in speech recognition often
depends heavily upon a language model to provide a prior probability
over likely word sequences. 

To integrate language model information during decoding,
\citet{Graves2014} use their CTC-trained neural network to rescore a
lattice or n-best hypothesis list generated by a state-of-the-art
HMM-based system. This introduces a potentially confounding factor
because an n-best list constrains the set of possible transcriptions
significantly. Additionally, it results in an overall system which
still relies on HMM speech recognition infrastructure to achieve the
final results. In contrast, we present \emph{first-pass} decoding
results which use a neural network and language model to decode from
scratch, rather than re-ranking an existing set of hypotheses.

We describe a decoding algorithm which directly integrates a language
model with CTC-trained neural networks to search through the space of
possible word sequences. Our first-pass decoding algorithm enables
CTC-trained models to benefit from a language model without relying on
an existing HMM-based system to generate a word lattice. This removes
the lingering dependence on HMM-centric speech recognition toolkits
and enables us to achieve fairly competitive WER results with only a
neural network and $n$-gram language model.

Deep neural networks (DNNs) are the most widely used neural network
architecture for speech recognition \citep{Hinton2012}. DNNs are a
fairly generic architecture for classification and regression
problems. In HMM-based LVCSR systems, DNNs act as acoustic models by
predicting the HMM's hidden state given the acoustic input for a point
in time. However, in such HMM-DNN systems the temporal reasoning about
an output sequence takes place within the HMM rather than the neural
network. CTC training of neural networks forces the network to model
output sequence dependencies rather than reasoning about single time
frames independently from others. To better handle such temporal
dependencies previous work with CTC used long short term memory (LSTM)
networks. LSTM is a neural network architecture was originally
designed to prevent the vanishing gradient problem of sigmoidal DNNs
or temporally recurrent deep neural networks (RDNNs) \citep{Hochreiter1997}.

Our work uses RDNNs instead of LSTMs as a neural network architecture. RDNNs
are simpler overall, because there are only dense weight matrix connections
between subsequent layers. This simpler architecture is more amenable to
graphics processing unit (GPU) computing which can significantly reduce
training times. Recent work shows that with rectifier nonlinearities DNNs can
perform well in DNN-HMM systems without suffering from vanishing gradient
problems during optimization \citep{Dahl2013,Zeiler2013,Maas2013}. This makes us
hopeful that RDNNs with rectifier nonlinearities may be able to perform
comparably to LSTMs which are specially engineered to avoid vanishing
gradients.

\section{Model}\label{sec:model}
We train neural networks using the CTC loss function to do maximum
likelihood training of letter sequences given acoustic features as
input. We consider a single utterance as a training example consisting
of an acoustic feature matrix $X$ and word transcription $W$. The CTC
objective function maximizes the log probability $\log p(W;X)$. We
reserve a full exposition of the loss function here because our
formulation follows exactly the previous work on using CTC to predict
the characters of an utterance transcription \citep{Graves2014,Graves2006}. 

\subsection{Deep Neural Networks}\label{sec:model:dnn}
With  the loss  function  fixed we  must  next define  how we  compute
$p(c|x_t)$,  the  predicted distribution  over  output characters  $c$
given  the audio  features  $x_t$  at time  $t$.  While many  function
approximators are possible for this  task, we choose as our most basic
model a DNN. A DNN computes the distribution $p(c|x_t)$ using a series
of hidden  layers followed by an  output layer. Given  an input vector
$x_t$ the first hidden layer activations are a vector computed as,
\begin{align}
  h^{(1)} = \sigma(W^{(1)T} x_t + b^{(1)}).
\label{eqn:hidden_input}
\end{align}
The matrix $W^{(1)}$ and vector $b^{(1)}$ are the weight matrix and
bias vector for the layer. The function $\sigma(\cdot)$ is a
point-wise nonlinearity. We use rectifier nonlinearities and thus
choose, $\sigma(z) = \max (z, 0)$.

DNNs can have arbitrarily many hidden layers. After the first hidden
layer, the hidden activations $h^{(i)}$ for layer $i$ are computed as,
\begin{align}
  h^{(i)} = \sigma(W^{(i)T} h^{(i-1)} + b^{(i)}).
\label{eqn:hidden_hidden}
\end{align}

To obtain a proper distribution over the set of possible characters
$c$ the final layer of the network is a \emph{softmax} output layer of the form,

\begin{align}
  p(c=c_k|x_t) = \frac{\exp(- ( W^{(s)T}_k h^{(i-1)} + b^{(s)}_k))}
  {\sum_j \exp(- ( W^{(s)T}_j h^{(i-1)} + b^{(s)}_j))},
\label{eqn:softmax}
\end{align}
where $W^{(s)}_k$ is the $k$'th column of the output weight matrix
$W^{(s)}$ and $b^{(s)}_k$ is a scalar bias term. 

We can compute a subgradient for all parameters of the DNN given a
training example and thus utilize gradient-based optimization
techniques. Note that this same DNN formulation is commonly used in
DNN-HMM models to predict a distribution over senones instead of
characters.

\subsection{Recurrent Deep Neural Networks}\label{sec:model:rdnn}
A transcription $W$ has many temporal dependencies which a DNN may not
sufficiently capture. At each timestep $t$ the DNN computes its output
using only the input features $x_t$, ignoring previous hidden
representations and output distributions. To enable better modeling of
the temporal dependencies present in a problem, we use a RDNN. In a
RDNN we select one hidden layer $j$ to have a temporally recurrent
weight matrix $W^{(f)}$ and compute the layer's hidden activations as,
\begin{align}
  h^{(j)}_t = \sigma(W^{(j)T} h^{(j-1)}_t +  W^{(f)T} h^{(j)}_{t-1} + b^{(j)}).
\label{eqn:hidden_rdnn}
\end{align}
Note that we now make the distinction $h^{(j)}_t$ for the hidden
activation vector of layer $j$ at timestep $t$ since it now depends
upon the activation vector of layer $j$ at time $t-1$.

When working with RDNNs, we found it important to use a modified
version of the rectifier nonlinearity. This modified function selects
$\sigma(z) = \min( \max (z, 0), 20)$ which clips large activations to
prevent divergence during network training. Setting the maximum
allowed activation to 20 results in the clipped rectifier acting as a
normal rectifier function in all but the most extreme cases.

Aside from these changes, computations for a RDNN are the same as
those in a DNN as described in \ref{sec:model:dnn}. Like the DNN, we
can compute a subgradient for a RDNN using a method sometimes called
backpropagation through time. In our experiments we always compute the
gradient completely through time rather than truncating to obtain an
approximate subgradient.

\subsection{Bi-Directional Recurrent Deep Neural Networks}\label{sec:model:brdnn}

While forward recurrent connections reflect the temporal nature of the 
audio input, a perhaps more powerful sequence transduction model is a
BRDNN, which maintains state both forwards and backwards in time.
Such a model can integrate information from the entire temporal extent
of the input features when making each prediction.
We extend the RDNN to form a BRDNN by again choosing a
temporally recurrent layer $j$.  The BRDNN creates both a forward and
backward intermediate hidden representation which we call $h_t^{(f)}$
and $h_t^{(b)}$ respectively. 
We use the
temporal weight matrices $W^{(f)}$ and $W^{(b)}$ to propagate
$h_t^{(f)}$ forward in time and $h_t^{(b)}$ backward in time
respectively.
We update the forward and backward components via the equations,
\begin{align}
\begin{split}
  &h^{(f)}_t = \sigma(W^{(j)T} h^{(j-1)}_t +  W^{(f)T} h^{(f)}_{t-1} + b^{(j)}), \\
  &h^{(b)}_t = \sigma(W^{(j)T} h^{(j-1)}_t +  W^{(b)T} h^{(b)}_{t+1} + b^{(j)}).
\end{split}
\label{eqn:hidden_brdnn}
\end{align}

Note that the recurrent forward and backward hidden representations
are computed entirely independently from each other. As with the RDNN
we use the modified nonlinearity function 
$\sigma(z) = \min( \max (z,0), 20)$. 
To obtain the final representation $h^{(j)}_t$ for the
layer we sum the two temporally recurrent components,
\begin{align}
\begin{split}
  h^{(j)}_t = h^{(f)}_t + h^{(b)}_t.
\end{split}
\label{eqn:hidden_brdnn_sum}
\end{align}

Aside from this change to the recurrent layer the BRDNN computes its
output using the same equations as the RDNN. As for other models, we
can compute a subgradient for the BRDNN directly to perform
gradient-based optimization.

\section{Decoding}\label{sec:decoding}


Assuming an input of length $T$, the output of the neural network will be $p(c;
x_t)$ for $t = 1,\ldots,T$. Again, $p(c; x_t)$ is a distribution over possible
characters in the alphabet $\Sigma$, which includes the blank symbol, given
audio input $x_t$. In order to recover a character string from the output of
the neural network, as a first approximation, we take the argmax at each time
step. Let $S = (s_1,\ldots,s_T)$ be the character sequence where $s_t =
\argmax_{c \in \Sigma} p(c; x_t)$. The sequence $S$ is mapped to a
transcription by collapsing repeat characters and removing blanks. This gives a
sequence which can be scored against the reference transcription using both CER
and WER.

This first approximation lacks the ability to include the constraint of
either a lexicon or a language model. We propose a generic algorithm which is
capable of incorporating such constraints. Taking $X$ to be the acoustic input
of time $T$, we seek a transcription $W$ which maximizes the probability,
\begin{equation}
  \label{eq:joint} p_{\text{net}}(W ; X) p_{\text{lm}}(W).  
\end{equation} 
Here the overall probability of the transcription is modeled as the product of
two factors: $p_{\text{net}}$ given by the network and $p_{\text{lm}}$ given by a language
model prior. In practice the prior $p_{\text{lm}}(W)$, when given by an $n$-gram
language model, is too constraining and thus we down-weight it and include a
word insertion penalty (or bonus) as
\begin{equation}\label{eq:amlm}
    p_{\text{net}}(W ; X) p_{\text{lm}}(W)^\alpha |W|^\beta.
\end{equation}
Alogrithm \ref{alg:decode} attempts to find a word string $W$ which maximizes equation
\ref{eq:amlm}.
\begin{algorithm}[bt]
  \caption{Prefix Beam Search: The algorithm initializes the previous set of
prefixes $A_\text{prev}$ to the empty string. For each time step and every
prefix $\ell$ currently in $A_\text{prev}$, we propose adding a character from the
alphabet $\Sigma$ to the prefix. If the character is a blank, we do not extend
the prefix. If the character is a space, we incorporate the language model
constraint. Otherwise we extend the prefix and incorporate the output of the
network. All new active prefixes are added to $A_\text{next}$. We then set
$A_\text{prev}$ to include only the $k$ most probable prefixes of $A_\text{next}$.
The output is the $1$ most probable transcript, although the this can easily
be extended to return an $n$-best list.}\label{alg:decode}
  \begin{algorithmic}
    \State $p_b(\emptyset; x_{1:0}) \gets 1,\; p_{nb}(\emptyset; x_{1:0}) \gets 0$
    \State $A_{\text{prev}} \gets \{\emptyset\}$ 
    \For {$t = 1, \ldots, T$}
      \State $A_{\text{next}} \gets \{\}$
      \For {$\ell \; \text{\bf in} \; A_{\text{prev}}$}
	\For {$c \; \text{\bf in} \; \Sigma$}
	  \If {$c = \text{blank}$}
	    \State $p_b(\ell; x_{1:t}) \gets p(\text{blank}; x_t) 
		(p_b(\ell ; x_{1:t-1}) + p_{nb}(\ell ; x_{1:t-1}))$
	    \State $\mbox{add } \ell \mbox{ to } A_{\text{next}}$
	  \Else
	    \State $\ell^+ \gets \mbox{concatenate }\ell \mbox{ and } c$
	    \If {$c = \ell_\text{end}$}
	      \State $p_{nb}(\ell^+; x_{1:t}) \gets p(c; x_t)p_b(\ell ; x_{1:t-1})$
	      \State $p_{nb}(\ell; x_{1:t}) \gets p(c; x_t)p_b(\ell ; x_{1:t-1})$
	    \ElsIf {$c = \text{space}$}
	    \State $p_{nb}(\ell^+; x_{1:t}) \gets p(W(\ell^+)| W(\ell))^\alpha 
		p(c; x_t)(p_b(\ell ; x_{1:t-1}) + p_{nb}(\ell ; x_{1:t-1}))$
	    \Else
	      \State $p_{nb}(\ell^+; x_{1:t}) \gets p(c; x_t)
		(p_b(\ell; x_{1:t-1}) + p_{nb}(\ell; x_{1:t-1}))$
	    \EndIf
	    \If {$\ell^+ \; \text{\bf not in} \; A_\text{prev}$}
	      \State $p_{b}(\ell^+; x_{1:t}) \gets p(\text{blank}; x_t)
		(p_b(\ell^+; x_{1:t-1}) + p_{nb}(\ell^+; x_{1:t-1}))$
	      \State $p_{nb}(\ell^+; x_{1:t}) \gets p(c; x_t)p_{nb}(\ell^+; x_{1:t-1})$
	    \EndIf
	    \State $\mbox{add } \ell^+ \mbox{ to } A_{\text{next}}$
	  \EndIf
	\EndFor
      \EndFor
      \State $A_{\text{prev}} \gets k 
	\text{ most probable prefixes in } A_{\text{next}}$
    \EndFor
    \State \Return $1 \text{ most probable prefix in } A_{\text{prev}}$
  \end{algorithmic}
\end{algorithm}
The algorithm maintains two separate probabilities for each
prefix, $p_{b}(\ell; x_{1:t})$ and $p_{nb}(\ell; x_{1:t})$. Respectively, these
are the probability of the prefix $\ell$ ending in blank or not ending in blank
given the first $t$ time steps of the audio input $X$.

The sets $A_{\text{prev}}$ and $A_{\text{next}}$ maintain a list of active
prefixes at the previous time step and proposed prefixes at the next time step
respectively. Note that the size of $A_{\text{prev}}$ is never larger than the
beam width $k$. The overall probability of a prefix is the product of a word
insertion term and the sum of the blank and non-blank ending probabilities,
\begin{equation}\label{eq:prefixprob}
  p(\ell; x_{1:t}) = (p_b(\ell; x_{1:t}) + p_{nb}(\ell; x_{1:t})) |W(\ell)|^\beta,
\end{equation}
where $W(\ell)$ is the set of words in the sequence $\ell$. When taking
the $k$ most probable prefixes of $A_{\text{next}}$, we sort each prefix using
the probability given by equation \ref{eq:prefixprob}.

The variable $\ell_{\text{end}}$ is the last character in the label sequence
$\ell$. The function $W(\cdot)$, which converts $\ell$ into a string of words,
segments the sequence $\ell$ at each space character and truncates any
characters trailing the last space. 

We incorporate a lexicon or language model constraint by including the
probability $p(W(\ell^+) | W(\ell))$ whenever the algorithm proposes appending
a space character to $\ell$. By setting $p(W(\ell^+) | W(\ell))$ to $1$ if the
last word of $W(\ell^+)$ is in the lexicon and $0$ otherwise, the probability
acts as a constraint forcing all character strings $\ell$ to consist of only
words in the lexicon.  Furthermore, $p(W(\ell^+) | W(\ell))$ can represent a
$n$-gram language model by considering only the last $n-1$ words in $W(\ell)$.

\section{Experiments}\label{sec:experiments}
We evaluate our approach on the 81 hour Wall Street Journal (WSJ) news
article dictation corpus (available in the LDC catalog as LDC94S13B
and LDC93S6B). Our training set consists of 81 hours of speech from
37,318 utterances. The basic preparation of transforming the
LDC-released corpora into training and test subsets follows the Kaldi
speech recognition toolkit's s5 recipe \citep{Povey2011}. However, we
did not apply much of the text normalization used to prepare
transcripts for training an HMM system. Instead we simply drop
unnecessary transcript notes like lexical stress, keeping transcribed
word fragments and acronym punctuation marks. We can safely discard
much of this normalization because our approach does not rely on a
lexicon or pronunciation dictionary, which cause problems especially
for word fragments. Our language models are the standard models
released with the WSJ corpus without lexical expansion. We used the
`dev93' evaluation subset as a development set and report final test
set performance on the `eval92' evaluation subset. Both subsets use
the same 20k word vocabulary. The language model used for decoding is
constrained to this same 20k word vocabulary.

The input audio was converted into log-Mel filterbank features with 23
frequency bins. A context window of +/- 10 frames were concatenated to
form a final input vector of size 483. We did not perform additional
feature preprocessing or feature-space speaker adaptation.  Our output
alphabet consists of 32 classes, namely the blank symbol ``\_'', 26 letters, 3
punctuation marks (apostrophe, ., and -) as well as tokens for noise
and space. 

\subsection{First-Pass Decoding with a Language Model}\label{sec:experiments:decoding}
We trained a BRDNN with 5 hidden layers, all with 1824 hidden units,
for a total of 20.9M free parameters. The third hidden layer of the
network has recurrent connections. Weights in the network are
initialized from a uniform random distribution scaled by the weight
matrix's input and output layer size \citep{Glorot2011}. We use the
Nesterov accelerated gradient optimization algorithm as described in
\citet{Sutskever2013} with initial learning rate $10^{-5}$, and
maximum momentum 0.95. After each full pass through the training set
we divide the learning rate by 1.2 to ensure the overall learning
rate decreases over time. We train the network for a total of 20 
passes over the training set, which takes about 96 hours using our
Python GPU implementation. For decoding with prefix search we use a beam size
of 200 and cross-validate with a held-out set to find a good setting of the
parameters $\alpha$ and $\beta$.  Table~\ref{tab:res_decode} shows word and character
error rates for multiple approaches to decoding with this trained BRDNN.

\begin{table*}[bt]
\caption{Word error rate (WER) and character error rate (CER) results
  from a BDRNN trained with the CTC loss function.  As a
  baseline (No LM) we decode by choosing the most likely label at each
  timestep and performing standard collapsing as done in CTC
  training. We compare this baseline against our modified
  prefix-search decoder using a dictionary constraint and bigram
  language model.}
\label{tab:res_decode}
\begin{center}
\begin{small}
\begin{tabular}{lrr}
\toprule
Model & CER & WER\\
\midrule
No LM & 10.0 & 35.8\\
Dictionary LM & 8.5 & 24.4\\
Bigram LM & 5.7 & 14.1\\
\bottomrule
\end{tabular}
\end{small}
\end{center}
\end{table*}

Without any sort of language constraint WER is quite high, despite the
fairly low CER. This is consistent with our observation that many
mistakes at the character level occur when a word appears mostly
correct but does not conform to the highly irregular orthography of
English. Prefix-search decoding using the 20k word vocabulary as a
prior over possible character sequences results in a substantial
WER improvement, but changes the CER relatively little. Comparing the
CERs of the no LM and dictionary LM approaches again demonstrates
that without an LM the characters are mostly correct but are
distributed across many words which increases WER. A large relative
drop in both CER and WER occur when we decode with a bigram
LM. Performance of the bigram LM model demonstrates that CTC-trained
systems can attain competitive error rates without relying on a
lattice or n-best list generated by an existing speech system.
\subsection{The Effect of Recurrent Connections}\label{sec:experiments:recurrence}
Previous experiments with DNN-HMM systems found minimal benefits from
recurrent connections in DNN acoustic models. It is natural to wonder
whether recurrence, and especially bi-directional recurrence, is an
essential aspect of our architecture. To evaluate the impact of
recurrent connections we compare the train and test CERs of DNN, RDNN,
and BRDNN models while roughly controlling for the total number of
free parameters in the model. Table~\ref{tab:res_recurrence} shows the
results for each type of architecture. 

\begin{table*}[bt]
\caption{Train and test set character error rate (CER) results for a
  deep neural network (DNN) without recurrence, recurrent deep neural
  network with forward temporal connections (RDNN), and a
  bi-directional recurrent deep neural network (BRDNN). All models
  have 5 hidden layers. The DNN and RDNN both have 2,048 hidden units
  in each hidden layer while the BRDNN has 1,824 hidden units per
  hidden layer to keep its total number of free parameters similar to
  the other models. For all models we choose the most likely character
  at each timestep and apply CTC collapsing to obtain a
  character-level transcript hypothesis.}
\label{tab:res_recurrence}
\begin{center}
\begin{small}
\begin{tabular}{lrrr}
\toprule
Model & Parameters (M) & Train CER & Test CER \\
\midrule
DNN & 16.8 & 3.8 & 22.3\\
RDNN & 22.0 & 4.2 & 13.5\\
BRDNN & 20.9 & 2.8 & 10.7\\
\bottomrule
\end{tabular}
\end{small}
\end{center}
\end{table*}

Both variants of recurrent models show substantial test set CER
improvements over the non-recurrent DNN model. Note that we report
performance for a DNN of only 16.8M total parameters which is smaller
than the total number of parameters used in both the RDNN and BRDNN
models. We found that larger DNNs performed worse on the test set,
suggesting that DNNs may be more prone to over-fitting for this
task. Although the BRDNN has fewer parameters than the RDNN it
performs better on both the training and test sets. Again this
suggests that the architecture itself drives improved performance
rather than the total number of free parameters. Conversely, because the gap
between bi-directional recurrence and single recurrence is small relative to a
non-recurrent DNN, on-line speech recognition using a singly recurrent network
may be feasible without overly damaging performance.


\section{Conclusion}\label{sec:conclusion}
We presented a decoding algorithm which enables first-pass LVCSR with
a language model for CTC-trained neural networks. This decoding
approach removes the lingering dependence on HMM-based systems found
in previous work. Furthermore, first-pass decoding demonstrates the
capabilities of a CTC-trained system without the confounding factor of
potential effects from pruning the search space via a provided
lattice.  While our results do not outperform the best HMM-based
systems on the WSJ corpus, they demonstrate the promise of CTC-based
speech recognition systems.

Our experiments with BRDNN further simplify the infrastructure needed
to create CTC-based speech recognition systems. The BRDNN is overall a
less complex architecture than LSTMs and can relatively easily be made
to run on GPUs since large matrix multiplications dominate the
computation. However, our experiments suggest that recurrent
connections are critical for good performance. Bi-directional
recurrence helps beyond single direction recurrence but could be
sacrificed in cases that require low-latency, online speech
recognition. Taken together with previous work on CTC-based LVCSR, we
believe there is an exciting path forward for high quality LVCSR
without the complexity of HMM-based infrastructure.

\small{
\bibliography{audio}
\bibliographystyle{icml2014}
}

\end{document}